\documentclass[10pt,a4paper,twocolumn]{article}

\usepackage{lineno}
\usepackage{authblk}
\usepackage{graphicx}
\usepackage[explicit]{titlesec}
\usepackage[labelfont=bf,labelsep=endash,font=footnotesize]{caption}
\usepackage{tabu}
\usepackage{xcolor}
\usepackage{subcaption}
\usepackage{tabularx}
\usepackage[hidelinks]{hyperref}
\usepackage[hang]{footmisc}
\usepackage[normalem]{ulem}
\usepackage[top=2.5cm, bottom=2.8cm, left=1.5cm, right=1.5cm]{geometry}
\usepackage{abstract}
\usepackage{mathtools}
\usepackage{balance}
\usepackage{wrapfig}
\usepackage{color,soul}

\usepackage{fancyhdr}

\usepackage{comment}
\usepackage{tikz}
\usepackage{multirow}
\usepackage{multicol}
\usepackage{varwidth}
\usepackage{array}
\usepackage{makecell} 
\setcellgapes{5pt}
\usepackage{graphicx}
\usepackage{tabularx}

\makeatletter
\renewcommand\AB@affilsepx{, \protect\Affilfont}
\makeatother

\providecommand{\keywords}[1]{\textbf{Keywords}\ \ \textendash\ \   #1}

\titleformat{\section}{\large\bfseries}{\thesection.}{1em}{\MakeUppercase{#1}}
\titlespacing*{\section}{0pt}{12pt}{6pt}

\titleformat{\subsection}{\large}{\thesubsection}{1em}{#1}
\titlespacing*{\subsection}{0pt}{12pt}{6pt}

\titleformat{\subsubsection}{\large\itshape}{\thesubsubsection}{1em}{#1}
\titlespacing*{\subsubsection}{0pt}{12pt}{6pt}

\newcommand{\ITUurl}[1]{\textcolor{blue}{\urlstyle{same}\url{#1}}}

\newcommand{\ITUpar}{\vspace{8pt}\par}

\renewenvironment{abstract}
               {\list{}{
               \setlength{\rightmargin}{0mm}
               \setlength{\leftmargin}{0mm}
               \vspace{-0.25in}
                \item[\textit{\textbf{\hspace{22pt}Abstract  }}  \textendash]\relax}}
               {\endlist}

\def\starttable{\vspace{6pt}\begin{table}[ht]\center}
\def\startfigure{\vspace{6pt}\begin{figure}[ht]\center}

\makeatletter
\def\tagform@#1{\maketag@@@{\ignorespaces#1\unskip\@@italiccorr}}
\makeatother

\title{\large{\textbf{\uppercase{Agnostic Learning for Packing Machine Stoppage Prediction in Smart Factories}}}}

\author[1]{\normalsize{Gabriel Filios}}
\author[2]{\normalsize{Ioannis Katsidimas}}
\author[3]{\normalsize{Sotiris Nikoletseas}}
\author[4]{\normalsize{Stefanos H. Panagiotou}}
\author[5]{\normalsize{Theofanis P. Raptis}}

\affil[1,2,3,4]{\normalsize{Department of Computer Engineering and Informatics, University of Patras, Greece and Computer Technology Institute and Press ``Diophantus'', Patras, Greece}}\ITUpar
\affil[5]{\normalsize{Institute of Informatics and Telematics, National Research Council (IIT-CNR), Pisa, Italy}}

\date{\vspace{-12pt}{\small NOTE: Corresponding author: Stefanos Panagiotou, spanagiotou@ceid.upatras.gr} \\  \endgraf\rule{\textwidth}{1pt}}

\fancypagestyle{FirstPage}{
\lfoot{©International Telecommunication Union, 2022. CC BY-NC-ND 3.0 IGO. \\\emph{ITU Journal on Future and Evolving Technologies, Volume 3, Issue 3, Pages 793-807, December 2022.} \href{https://doi.org/10.52953/LEDZ3942}{https://doi.org/10.52953/LEDZ3942}}
}

\begin{document}


\twocolumn[

\begin{@twocolumnfalse}
\maketitle

\begin{abstract}
The cyber–physical convergence is opening up new business opportunities for industrial operators. The need for deep integration of the cyber and the physical worlds establishes a rich business agenda towards consolidating new system and network engineering approaches. This revolution would not be possible without the rich and heterogeneous sources of data, as well as the ability of their intelligent exploitation, mainly due to the fact that data will serve as a fundamental resource to promote Industry 4.0. One of the most fruitful research and practice areas emerging from this data-rich, cyber-physical, smart factory environment is the data-driven process monitoring field, which applies machine learning methodologies to enable predictive maintenance applications. In this paper, we examine popular  time series forecasting techniques as well as supervised machine learning algorithms in the applied context of Industry 4.0, by transforming and preprocessing the  historical industrial dataset of a packing machine's operational state recordings (real data coming from the production line of a manufacturing plant from the food and beverage domain).  In our methodology, we use only a single signal concerning the machine's operational status to make our predictions, without considering other operational variables or fault and warning signals, hence its characterization as ``agnostic''. In this respect, the results demonstrate that the adopted methods achieve a quite promising performance on three targeted use cases.
\end{abstract}

\ITUpar


\ITUpar
\ITUpar

\end{@twocolumnfalse}
]

\section{Introduction} 
\label{sec:intro}
\begin{figure*}
\caption{Methodology Workflow}
\includegraphics[width=\textwidth,height=10.5cm]{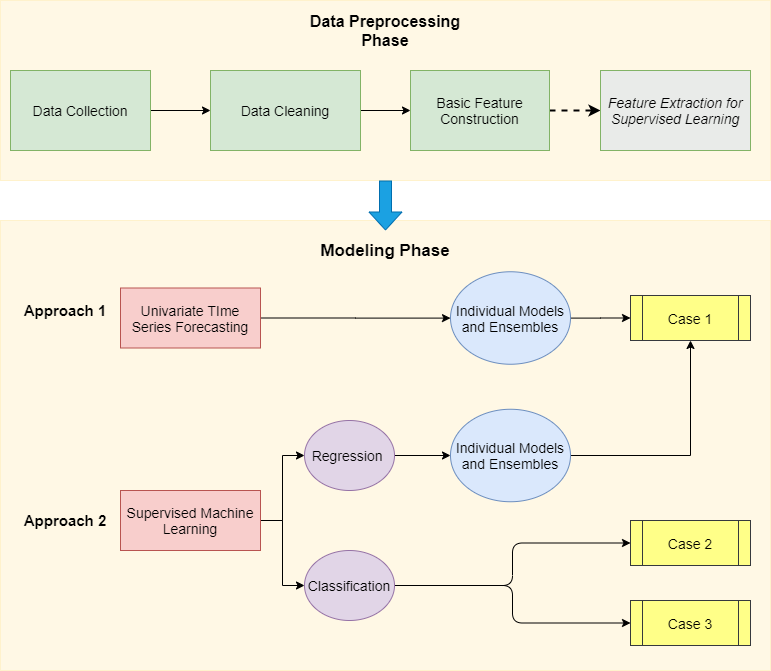}
\label{fig:methodology-workflow}
\end{figure*}

\thispagestyle{FirstPage}
Cyber-physical systems are an inevitable outcome of the fourth industrial revolution (also coined as Industry 4.0). Embedded computing, Internet communication, and ubiquitous control have now become fundamental components of modern engineered products and their manufacturing processes \cite{BarnardFeeney2017}. The cyber–physical convergence is opening up new business opportunities for industrial operators. The vision of a virtual world, that is overlaid on the physical world to continuously monitor it and take intelligent actions to adapt the cyber world to industrial needs, is part of an emerging trend of in pervasive and mobile computing  \cite{CONTI20122}. The need for deep integration of those interrelated worlds opens up a rich business agenda towards establishing a new system and network engineering that is both physical and virtual \cite{cps2012}. Consequently, the industrial cyber-physical systems employment is expected to revolutionize the way enterprises conduct their business from a holistic viewpoint, i.e., from shop-floor to business interactions, from suppliers to consumers, and from design to testing across the entire product and service lifecycle \cite{7883993}.
\ITUpar
This revolution would not be possible without the rich and heterogeneous sources of data, as well as the ability of their intelligent exploitation; mainly due to the fact that data will serve as a fundamental resource to promote Industry 4.0 from machine automation to information extraction and then to knowledge discovery \cite{8764545}. Smart factories already operate using sophisticated sensors, actuators and communication technologies. Internet of Things (IoT) devices are not seen any more as ``dumb things'' generating individually a few bytes, but as industrial devices, generating data of variable size and significant importance, still operated via batteries to make them more flexible and cheap to assemble, install and manage \cite{9039732}. Smart  factories perform adaptive responses by continuously monitoring and extracting information from physical objects (e.g., machines, work pieces, robotic elements, etc.) and production processes \cite{10.1007/978-3-319-65151-4_10}. In this setting, large amounts of data are generated and collected, requiring advanced big data processing methodologies to build an integrated environment in which the production processes can be mirrored transparently and administrated in a more efficient way \cite{8085101}. One of the most fruitful research and practice areas emerging from this data-rich, cyber-physical, smart factory environment is the data-driven process monitoring field, which applies multivariate statistical methods to enable prognostics, diagnostics and fault detection for industrial process operations and production results \cite{QIN2012220}. By applying big data analytics, it is possible to find interpretive results for strategic decision-making, providing novel insights which lead to significant production improvements, such as, maintenance cost decrease, early fault detection, machine stoppage prediction, spare parts inventory reduction, increased production, improvement in operator safety and repair verification \cite{CARVALHO2019106024}.
\ITUpar
Novel developments in specialized fields of information and communication technologies and availability of easy-to-use, often freely available software tools and off-the-shelf hardware components, offer great potential to transform the smart factory domain and their impact on the smart factory data pools effectively. One of the most trending developments is in the area of Machine Learning (ML). The utilization of ML is motivated by its enhanced capabilities to spare resources, machining time and energy, and its improved operational capacity where traditional methods have reached their limits \cite{weichert2019}. However, the field of ML is highly diverse and many different algorithms, theories, and methods are available. For many industrial operators, this represents a barrier regarding the adoption of powerful ML tools and thus may block the usage of the huge amounts of data which are more and more becoming available \cite{doi:10.1080/21693277.2016.1192517}. Additionally, industrial technology providers have chosen to deploy standalone systems which act as black boxes from which, in most cases, vital data is not possible to be acquired. Furthermore, it is also worth noting that, in many cases, the lack of qualified infrastructure, due to cost-related reasons such as the high commission fees of third party providers, is also a common problem.
\ITUpar
Based on the aforementioned assumptions, it is evident that there are large and complex production lines which suffer from the absence of the necessary, sophisticated prediction infrastructure and, therefore, from the inability to receive the corresponding data and maintain a rich data pool. In this paper, we argue that even those can become able to establish a reliable and efficient prediction ability, in order to activate the required predictive maintenance (PdM) mechanisms. Specifically, we aim to predict a packing machine's stoppages on the following three targeted use cases: 
\begin{enumerate}
    \item Forecasting of the total daily duration of stoppages in the near future 
    \item Prediction of whether the packing machine will be stopped for more than 10 minutes in the next hour (later mentioned as ``Minor  Stoppage  Duration  Exceeding" case)
    \item Prediction of whether a specific type of stoppage (breakdown event) will occur in the next hour (later mentioned as ``Breakdown Occurrence" case)
\end{enumerate}
In order to apply our proposed methodologies for these cases, we are taking advantage of univariate time series data coming from a production line machine of a large manufacturing plant from the food and beverage domain\footnote{A non-disclosure agreement prevents us from providing more information regarding the company, the plant, the equipment and the related data.}. We aim at using  the entire behavior of the machine as a single piece of data, or, in other words, we investigate whether a sequence of interruptions of a specific duration can potentially lead to the production line's stoppage in the near future.
\ITUpar
Specifically, we first transform the collected raw dataset with several essential and required data preprocessing procedures and then we leverage this transformed time series data with two core approaches. The first one, consists of univariate time series forecasting algorithms (Prophet, ARIMA, HWAMS, TBAT, N-BEATS) that are employed for forecasting the first aforementioned use case. We also produce forecasting ensembles of these models to examine if we can achieve higher accuracy using multiple combinations of individual models. The second approach leverages machine learning algorithms to model our time series data, reframing the latter into a feature-based dataset. We utilize tree-based machine learning regression algorithms (Decision Tree, XGBoost Regressor, Extra-Trees Regressor, AdaBoost, Gradient Boosting Regressor), both individually and as ensembles, in the same use case we apply the time series forecasting algorithms and compare their performance and results. Lastly, we investigate the next two aforementioned use cases as binary classification problems, using the Random Forest algorithm to make our predictions. 
\ITUpar
To make this methodology clear, we offer a high level visual representation of our overall workflow in Figure \ref{fig:methodology-workflow}.

\section{Related Work}
Recently, ML services in the context of Industry 4.0 and PdM are trending in the ICT and manufacturing fields. More and more applied research is conducted to come up with reliable and insightful predictive analytics workflows and solutions which can be adopted by industrial manufacturing companies. Here, we present some indicative research works which can be considered as the most relevant to our research. In order to narrow down the presentation, we focus on approaches which use intelligent techniques and take advantage of real data coming from actual industrial plants.
\ITUpar
The authors of \cite{GUTSCHI2019528}, present a data-driven approach for estimating the probability of machine breakdown during specified time interval in the future, in the context of discrete parts manufacturing systems. Specifically, they utilize data such as historical log messages, event logs and operational information from milling machines in order to estimate the Remaining Useful Life (RUL) by using the Random Forest machine learning algorithm. Results demonstrate that machine failures can be reliably predicted up to 168 hours in advance and that feature selection algorithms highly impact log-based predictive maintenance (PdM).
\ITUpar
Research work \cite{10.1007/978-3-030-20948-3_11} studies a predictive maintenance application in the White Goods/Home Appliances sector. The use case concerns a single production line that produces drums for dryers and the objective is to predict if the factory is going to have a breakdown on the next day, using legacy and operational data. Several machine learning algorithms are employed for modeling this binary classification problem such as Support Vector Machine (SVM), the multinomial Naïve Bayes classifier, k-NN, Decision Tree, Random Forest, and the Multilayer Perceptron (MLP), from which Decision Tree reaches  the highest accuracy score. Another interesting finding is the usage of an additional custom metadata feature in the modeling, calculated in the original dataset, that helped to increase the accuracy from  77.7\% to the final and best score of 95\%.
\ITUpar
In \cite{JAEC74}, the authors develop a Bayesian network for fault assessment of an electrical motor. Their proposed model is able to calculate  through inference the probability of rotor fault of an induction motor and  define the weakest branch in the structure of the Bayesian network that leads to failure by determining the probabilities of intermediate events. Their methodology is validated in a motor of a real industrial site, using its history file and the results of conducted diagnostics to determine the a priori probabilities of several potential fault causes and thus enabling the Bayesian network to define the most likely faults.
\ITUpar
The authors of \cite{Moleda2020}, examine a predictive maintenance use case, proposing a method for early detection of faults in boiler feed pumps. In particular, they utilize existing event data and  measurements acquired from SCADA systems of a coal fired power plant. The aim is to produce a model that can  detect deviations from the normal operation state based on regression and to check which events or failures can be detected by it. Their proposed algorithm consists of a bag of regression models and the experimental results indicate that it outperforms decision trees and multi-layer perceptron (MLP) classification algorithms.
\ITUpar

In research work \cite{pasteurAIzer}, a data-driven soft sensing methodology is presented for monitoring the quality of an industrial pasteurization process. The methodology is based on machine learning and benchmarks various algorithms such as Decision Tree, Ridge Regression, Extra Trees for estimating the temperature of products during the pasteurization process. The work studies a real beer pasteurization process in collaboration with Heineken's plant in Patras, Greece and the results demonstrate notable performance in temperature prediction accuracy, with average root mean square error (RMSE) of 1.85°C in the test sets.
\ITUpar
The authors of \cite{RUIZSARMIENTO2020103289} investigate the machinery involved in the production of high-quality steel sheets by predicting the degradation of the drums within the involved heating coilers. They incorporate valuable sources of information like  expert knowledge, configuration parameters and real time information coming from sensors mounted on the machines, in order to develop a predictive model based on a Discrete Bayesian Filter (DBF) that estimates and predicts the gradual degradation of such machinery. The DBF model reaches high success, compared with knowledge based  and classical machine learning models and appears to be robust against noisy and fluctuating data and integrate well expert knowledge with production data.
\ITUpar
In \cite{SUBRAMANIYAN2018533}, authors propose a data-driven algorithm to predict throughput bottlenecks in a real-world production system from an automotive production line. They employ an auto-regressive integrated moving average (ARIMA) method to predict the active run periods of the machine and integrate it with a data-driven active period technique to form a binary classification bottleneck prediction problem. Authors of \cite{8466309} present a methodology that utilizes process sensor data from operation periods to forecast possible equipment stoppages (or faults) of a specified industrial equipment, which is used in the anode production process for the aluminium industry. The classifiers employed were the Decision Tree, the Random Forest, the Gaussian/Bernoulli Naive Bayes and the Multilayer Perceptron including the Logistic Regression. The visualization of the features patterns and the simulation results show that a warning timeframe around 5-10 minutes before the incident occurs is a feasible goal.
\paragraph{\textbf{Contribution and motivation of our work}} 

Different to the aforementioned research works, and as an extended version of our previous work in \cite{dcos2020}, this paper includes the following novelties:
\begin{itemize}
    \item Investigation of a packing machine's stoppages in a production line of a manufacturing plant from the food and beverage domain.
    \item  Complex utilization of a single production scale machine signal in order to create a dataset that will allow the  modeling of  the machine's stoppage behaviour. The machine is located at the final stages of the packing manufacturing process and there are multiple machinery, human and material reasons and issues that appear earlier in the process and cause it to stop. However, we are not able to monitor these and we only monitor the signal describing its operational state, hence the characterization of our methodology  as ``agnostic", which contradicts approaches that apply ML techniques and require large pools of data coming from different sources.
    \item We not only attempt to investigate interesting, to industrial operators and the general literature, use cases but we also aim at testing the potential self-forecastability of this univariate time series, by predicting the machine's behaviour based on the past measurements of its operational state.
    \item Extraction of custom metadata features from the original signal data in order to enable the employment of machine learning classification and regression algorithms.
    \item Comparison of traditional forecasting and  machine learning regression models, as well as their generated ensembles, in the applied context of industry 4.0.
    \item The generated ensembles are proven to produce more accurate forecasts than simple, naive models.
\end{itemize}

The use cases investigated in this work are arguably very interesting, as the packing machine is considered to be the bottleneck of the production line. In this way, a potential successful predictive modeling can give shop-floor operators the opportunity to take proactive actions and scheduling of the production line, resulting in great cost reduction and higher plant efficiency and productivity.

\section{ Data Analysis and Preprocessing}\label{methodology} 

In this section we describe how  the raw data is collected, analysed and transformed in order to create a meaningful data representation for our modeling approaches. All of our proposed steps below are implemented using Python and Pandas, the powerful data analysis library \cite{mckinney-proc-scipy-2010}.

\subsection{Data Collection \& Outline}\label{data-outline}

As we have already mentioned, our factory application is related to the food and beverages sector. An IoT controller is installed to receive digital signals regarding the packing machine running state (\textit{RUN} and \textit{STOP} states) of the production line. In particular, the controller is a Raspberry Pi 3 Model B that is connected to the output of a Siemens SIMATIC S7-300 programmable logic controller (PLC) and receives and stores the data locally for further analysis. Data acquisition is implemented in a clear and transparent way without interfering in any way to the process, which may cause profit loss. More signals and information require great effort, as data are gathered in different data storage modules, which in many cases consider "black box" systems, the topology does not allow an easy wired data transfer, and current infrastructure do not offer many options on data transmission. In addition, the presence of many metal surfaces, motors and inverters cause a lot of noise and prevents the use of wireless communications, which in practice appear to have low reliability. Thus, working with current dataset was our only option.

\ITUpar
The dataset consists of irregular event-based time series, containing the stoppages and running states of an industrial packing machine which  alternate at arbitrary times and as a result  the spacing of the observation times is not constant. It is collected in a total period of one year, from October 2018 to October 2019 and contains 73576 entries from which  36775 are stoppages. Each time the machine’s operation state  changes either from \textit{RUN} to \textit{STOP} and vice  versa, it is logged as an event with three columns: 
\begin{itemize}
    \item the timestamp in which the state changed
    \item the binary value of the machine’s  state (1 for \textit{RUN} and 0 for \textit{STOP}) 
    \item the duration of the previous state in seconds by calculating their timestamp difference
\end{itemize}

Table~\ref{table:raw-data}, reflects a short snapshot of the raw data.

\begin{table}[!htbp]
\centering
\caption{Example of the collected raw data}
    \begin{tabular}{|c|c|c|} 
    \hline
    Timestamp  & State & Duration  \\
    \hline
    2019-05-30 14:44:09.298 & 1 & 2.141 \\
    \hline
    2019-05-30 14:44:12.866 & 0 & 3.567 \\
    \hline
    2019-05-30 14:44:24.994 & 1 & 12.128 \\
    \hline
    2019-05-30 14:44:30.127 & 0 & 5.133 \\
    \hline
    2019-05-30 14:45:15.596 & 1 & 45.469 \\
    \hline
    \end{tabular}
    
    \label{table:raw-data}
\end{table}

It is also  worth mentioning that industrial operators further discriminate the stoppages based on their duration, resulting in three stoppage categories: 

\begin{itemize}
\item \textit{Minor stoppages}, those that last between 10 seconds and 5 minutes
\item \textit{Breakdown stoppages}, those that last between 5 and 40  minutes
\item \textit{Major stoppages}, those that last more than 40 minutes
\end{itemize} 

In Tables ~\ref{table:working-statistics} and \ref{table:stoppages-statistics}, the packing machine's daily operational RUN time statistics in hour scale and count of stoppages per category are presented respectively for the time period of October 2018 to October 2019.

\begin{table}[!htbp]
\centering
   \caption {Daily RUN time statistics in hour scale}
    \begin{tabular}{|c|c|}
    \hline
    Metric & Value \\  \hline
    min & 2.1 \\ \hline 
    max & 20\\  \hline
    Daily Mean  & 14\\  \hline
    Daily Median & 14.5 \\  \hline
    Daily St.Deviation & 3.5\\  \hline
    \end{tabular}
    \label{table:working-statistics}
\end{table}

\begin{table}[!htbp]
\centering
\caption {Machine's stoppages count statistics}
    \begin{tabular}{|c|c|}
    \hline
    Count & Value \\  \hline
    Daily Mean (all categories) & 180 \\  \hline
    Minor & 32441\\  \hline
    Breakdown & 2545\\  \hline
    Major & 289 \\   \hline
    \end{tabular}
    \label{table:stoppages-statistics}
\end{table}

\subsection{Dataset Cleaning}
The first step of the data preparation process involves the application of some core  preprocessing steps in order to clean up the data from outliers and noisy, inconsistent  information that can decrease the predictive performance of the models. 

First of all, a small number of  “duplicates” (consecutive rows with the same value,  either 1 or 0) is removed from the dataset as it is important to have strictly alternating states. Afterwards, we need to assign the duration of each event in its own row. So, based on the aforementioned raw format, a shifting operation is applied in the duration column in order to set for each row the duration of it’s next row. A third step is to tidy up the data based on the normal operation days and hours of the machine (and the  whole production line in general). Any events that might have happened outside these (e.g after Saturday evenings, before the first shift of Mondays)  are considered as special events (e.g for testing or maintenance purposes)  and they are removed.

\subsection{Feature Engineering}
This is the most important step prior to applying our proposed  modeling approaches. Section \ref{subsubsec:basif-fe} describes the core methodology to create a meaningful dataset. This dataset is used to model univariate time series forecasting problems, as presented in Section \ref{forecasting-methods}. In Section \ref{feature-extraction} we build upon this dataset and extract additional features to be  used in our machine learning regression and classification approaches.

\subsubsection{Basic Data Preparation}\label{subsubsec:basif-fe}
It is evident that the described dataset in its raw format is not capable of providing useful information as input to a prediction model. Therefore,  following the cleaning and preprocessing  steps it is essential to add more representative information about the machine’s functional  patterns and we accomplish this by extracting features from the same data we have. 

One very common operation is to  “resample” (group)  the data in different  time intervals each time (e.g by 5, 10, 15, 30 minutes, hourly or daily resampling)  and calculate aggregation and statistical features such as  the sum, mean, standard deviation of the durations or the count of the stoppage and running events in each interval. 

It is worth mentioning that this resampling procedure  requires additional preparation of the aforementioned dataset in order to calculate accurately the above features. As an example, assuming the resampling by 5 minutes, an interval 09:05-09:10 and a stoppage occurring at 09:09 until 09:15 (6 minutes duration), the event must be splitted in two events. One occurring at 09:09 with duration of 1 minute and one occurring at 09:10 with duration of 5 minutes. In this way, the initial information of the duration is not altered but it is just distributed correctly between the intervals 09:05-09:10 and 09:10-09:15 for the feature calculations. 

\begin{figure}[!htbp]
\caption{\textit{Splitted resampling representation}}
\includegraphics[width=\columnwidth]{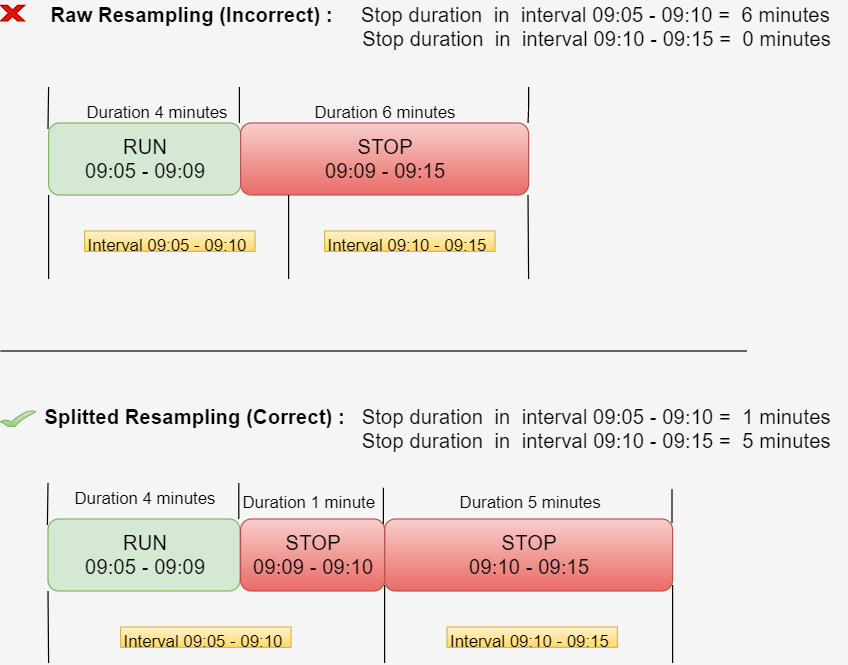}
\label{fig:resampling}
\end{figure}

In Figure \ref{fig:resampling}, a visual representation of this procedure is presented, while Table \ref{table:resampled-dataset} depicts a dataset example concerning ttotal duration (in seconds) and count of stoppages per hour.


\begin{table}[!htbp]
\caption{Example of an hourly resampled dataset of stoppages}
    \centering
    \begin{tabular}{|c|c|c|}
    \hline
    Timestamp  & Sum  & Count  \\
    \hline
    2019-05-30 14:00:00 & 1176.647	 & 14  \\
    \hline
    2019-05-30 15:00:00 & 1133.857	  & 23 \\
    \hline
    2019-05-30 16:00:00 & 44.975 &  2 \\
    \hline
    2019-05-30 17:00:00 & 818.693  & 7   \\
    \hline
    2019-05-30 18:00:00 & 298.273  & 5 \\
    \hline
    \end{tabular}
    
    \label{table:resampled-dataset}
\end{table}
\ITUpar
\subsubsection{Feature Extraction}\label{feature-extraction}

Supervised machine learning algorithms like the ones we use in our experiments depend on meaningful features, in order to correlate and map the input data with the output class of interest. The relevancy of features with the class to be predicted is a factor of high importance for the successful modeling and prediction performance of the algorithms. In our case, we generate a rich feature set by adding various features based on the process described in \ref{subsubsec:basif-fe}, and the knowledge of the industrial operators, reaching into a final number of 115 features.

The core feature list consists of the total duration (\textit{sum}) and the number (\textit{count}) of the occurred stoppages and run events in a given interval. Specifically, concerning the stoppages, we calculate sum and count features for each aforementioned stoppage category. Some performance indicators can be also  synthesized based on  \textit{Minor stoppages} as they are the most frequent type of stoppages and arguably the most important for analysing the machine's behaviour. For example, \textit{Mean Time Between Stops (MTBS)} and \textit{Mean Time Between Runs (MTBR)}, give an intuitive reflection on how frequently do stoppages and run events respectively occur, during the specified interval, and are calculated as follows :

\begin{equation} MTBS = \frac{sum(run)}{count (stops)}\end{equation}  \hspace{1cm} \begin{equation} MTBR = \frac{sum(stop)}{count (runs)} \end{equation}

Furthermore, based on the feature list above \textit{(sum and count of [minor\_stop, breakdown\_stop, major\_stop], sum and count of runs, MTBS, MTBR)}, we extract two distinct feature categories which are calculated for each feature of the former. The first one, consists of the so-called lag features, which are variables containing values from prior time intervals and particularly, our lags range from 1 to 5 prior time steps. For example, assuming an hourly input window interval and a specific row in the dataset corresponding to the data for the input interval between 09:00-10:00,  \textit{count\_stop} represents the count of stops of that hour and \textit{count\_stop\_t\_1}, \textit{count\_stop\_t\_2}  the count of stops occurred in the intervals 08:00-09:00 and  07:00-08:00, respectively. The second category is about rolling statistics features, where we calculate the moving arithmetic mean and standard deviation of the core features with their lag values.
\ITUpar
At last, we also  create the  \textit{time\_since\_[major\_stop, breakdown\_stop]} which is the time elapsed (in minutes) from the last occurrence of a  \textit{[major\_stop, breakdown\_stop]} until each input time interval. Besides this, we decompose the timestamp column of Table ~\ref{table:resampled-dataset} into single  date time features like \textit{day\_of\_week}, \textit{day\_of\_month}, \textit{week}, \textit{hour} for each input time interval. The purpose is to provide the  models  with an understandable representation of the time nature of the data so they can potentially catch and model any underlying time related patterns such as trends and seasonalities.

\section{Proposed Modeling Approaches}\label{sec:mod}

Our proposed modeling approaches are distinguished between two core categories. The first one, is univariate time series forecasting, which takes into account only one time dependent (numerical) variable and by modeling only  its historical values, it produces forecasts for future time points. The second one, is the supervised machine learning approach, which differentiates from the former by using features in order to model and map the input data to an output, the prediction target of interest.

\subsection{Univariate Time Series Forecasting}\label{forecasting-methods}

Univariate time series data is a collection of observations with chronological order. In particular, it consists of the independent variable, time, and just one dependent variable, which varies over time. Forecasting this kind of data means that past measurements are analysed by the algorithms to model the underlying pattern and behaviour of the series and make predictions for future points. Our use case concerns the forecasting of the daily total sum of minor stoppages, i.e, how much time will the investigated machine be in stop state per day in the future, due to minor stoppages. In particular, the dataset consists of the timestamp column, which ranges from January 2019 to August 2019 and  the \textit{sum\_minor\_stop} feature column. Daily resampling of the observations is chosen as it is an important indicator for industrial operators when monitoring the machine’s efficiency. It is not by coincidence that we examine this use case, as minor stoppages are the most frequent type of stoppages and are considered to be the bottleneck of the machine’s operation. Concerning the selected date period, in these dates the machine is more busy than any other period during the year, and thus, succeeding in generating insightful forecasts is vital to industrial operators. Model wise, we use the AtsPy time series library \cite{atspy} for the implementation of the models described below. After producing forecasts for each individual model, we create their ensembles as described in \ref{ensembles}.

\begin{itemize}
    \item \textit{ARIMA}, stands for Autoregressive Integrated Moving Average model, which is expressed as ARIMA(p,d,q). Parameters p, d, and q are integer values that decide the structure of the time series model; parameter p, q is the order of the AutoRegressive (AR) model and the Moving Average (MA) model respectively, while parameter d is the level of differencing applied to the data, if non stationarity is detected \cite{HO1998213,doi:10.1002/for.3980020104}.
    \item \textit{Prophet}, is a procedure for forecasting time series data based on an additive model where non-linear trends are fit with yearly, weekly, and daily seasonality, plus holiday effects. It is robust to missing data and shifts in the trend and typically handles outliers well \cite{taylor2018forecasting}.
    \item  \textit{TBAT}, is a method that accounts for  multiple seasonalities, using a combination of Fourier terms, exponential smoothing models, and Box-Cox transformations in a completely automated manner. TBATS also allows for the seasonal patterns to dynamically adjust over time \cite{doi:10.1198/jasa.2011.tm09771}.
    In the ensembles generation, TBATS  is a simple variation of TBAT, crafted in the AtsPy library for  using seasonal transformation and TBAT without seasonal.
    \item \textit{N-BEATS}, is a deep neural architecture based on backward and forward residual links and a very deep stack of fully-connected layers incorporated for solving univariate times series point forecasting problems \cite{oreshkin2019nbeats}.
    \item  \textit{HWAMS}, implements Holt Winter's exponential smoothing with additive trend and multiplicative seasonality \cite{HOLT20045}.
\end{itemize}

\subsection{Supervised Machine Learning}\label{ml-algorithms}

In this part we attempt to reframe the univariate time series forecasting approach as a supervised machine learning problem. The motivation for this feature-based time series modeling is to explore new valuable, for industrial operators, prediction use cases, such as the prediction  of a stoppage occurrence in the near future. Moreover, we can potentially overcome some limitations of the data, that univariate time series forecasting methods usually can not handle efficiently. For example, such limitations include the lack of massive historical data from a long time period to capture seasonality, and the fact that the past measurements of just the prediction target can not be enough to model it’s behavior, as it might be also dependent to other exogenous factors.

Considering these, we transform the univariate time series data into a feature-based dataset, that machine learning algorithms can take advantage of. To achieve this, the dataset must undertake a considerable transformation as there is no concept of input and output features in time series. A supervised learning framing of a time series means that the data needs to be splitted into multiple samples that the models can learn from, and generalize across. Each sample (row) must have both an input set of features (columns) and an output component, i.e,  the prediction target, also known as the class of the problem. We extract and generate the input set of features as presented in Section \ref{feature-extraction}. Regarding the prediction target, we describe our use case decisions accordingly in Section \ref{subsubsec:regression} and \ref{classification}. The model implementation and evaluation are made using Python and the Scikit-Learn library \cite{scikit-learn}.

\subsubsection{Regression Approach}\label{subsubsec:regression}

A machine learning regression approach considers the modeling of input data to predict numerical (integer or continuous) values. In this part, we leverage machine learning regression algorithms, both individually and as ensembles (see Section \ref{ensembles}), for the same use case and experiment details we applied the univariate time series forecasting models  (time period ranging from January 2019 to August 2019 and taking into account only information from minor stoppages). The dataset is constructed in a way so that the input features (see Section \ref{feature-extraction}) correspond to the previous day of  the prediction target. In other words, input features of each sample (row) are mapped with the next day's value of \textit{sum\_minor\_stop} target.

Model wise, we utilize the following tree-based regression algorithms.

\begin{itemize}
    \item \textit{Decision Tree}, a simple  regression tree  implementation.
    \item \textit{Extra Trees}, implement a meta estimator that fits a number of randomized decision trees (a.k.a. extra-trees) on various sub-samples of the dataset and uses averaging to improve the predictive accuracy and control over-fitting \cite{10.1007/s10994-006-6226-1}.
     \item \textit{Gradient Boosting Regressors (GBR)},   is a machine learning technique for regression problems, which produces a prediction model in the form of an ensemble of weak prediction models, typically decision trees. At each step, a new tree is trained against the negative gradient of the loss function, which is analogous to (or identical to, in the case of least-squares error) the residual error \cite{FRIEDMAN2002367}. 
    \item \textit{eXtreme Gradient Boosting (known as XGBoost and denoted with XGB in the ensembles)}, is a decision-tree-based ensemble Machine Learning algorithm that uses a gradient boosting framework. In particular, it is about a scalable and accurate implementation of gradient boosting machines and it has proven to push the limits of computing power for boosted trees algorithms as it was built and developed for the sole purpose of model performance and computational speed \cite{10.1145/2939672.2939785}.
   \item {AdaBoost}, an AdaBoost regressor is a meta-estimator that begins by fitting a regressor on the original dataset and then fits additional copies of the regressor on the same dataset but where the weights of instances are adjusted according to the error of the current prediction. As such, subsequent regressors focus more on difficult cases \cite{Drucker97improvingregressors, FREUND1997119}.
   
\end{itemize}

\subsubsection{Classification Approach}\label{classification}

In our machine learning classification approach, we investigate two distinct binary classification tasks. 
\begin{enumerate}
  
    \item Prediction of whether the packing machine will be stopped for more than 10 minutes in the next hour (``Minor  Stoppage  Duration  Exceeding" case)
    \item Prediction of whether a specific type of stoppage (breakdown event) will occur in the next hour (``Breakdown Occurrence" case)
\end{enumerate}
Both cases constitute a very important and interesting challenge. A potential accurate predictive modeling could warn industrial operators and in this way give them enough time to take preventive and corrective maintenance actions in the machine's production line or the investigated machine itself. As a result, this could lead to avoiding stoppages from occurring, and thus, to significant cost reductions for the unit. Regarding the dataset, in both cases, we calculate the aforementioned features of Section \ref{feature-extraction}, for an hourly input window and resampling interval. This means that we gather and aggregate the data per hour and we map each hour interval (row)  to the prediction target (class) for the next hour.

We use Random Forest classifier as the modeling algorithm for both cases. A random forest is a meta estimator that fits a number of decision tree classifiers on various sub-samples of the dataset and uses averaging to improve the predictive accuracy \cite{10.1023/A:1010933404324}. We chose this classifier as it is considered a highly accurate and robust method because of the number of decision trees participating in the process. It does not suffer from the overfitting problem, canceling out the biases by generating an internal unbiased estimate of the generalization error as the forest building progresses.

\subsection{Average Ensembles Generation}\label{ensembles}

Forecasting accuracy can potentially be improved by combining forecasts, produced by different algorithms \cite{GRAEFE201443,ARMSTRONG1989585}. Thus, we attempt such an ensemble approach as well, generating two distinct ensemble categories from  the aforementioned univariate forecasting and regression models, respectively. In particular, multiple combinations of the individuals models compose an ensemble by \textit{averaging} their predictions for each data point in the test data. 

We produce, in a brute force manner, all possible combinations of ensembles of different lengths (subsets) between the models, generating iteratively ensembles consisting of 
\begin{equation} 
i = 2...k   \hspace{0.2cm} individual \hspace{0.2cm} models
\end{equation} 
where $i$, is the number of individual models in each ensemble and  $k$, is the total number of individual models in each model category (forecasting and regression). 

Taking forecasting category as an example, which consists of five (5) individual models, we generate all possible ensembles that consist of 2, 3, 4 and 5 individual models respectively. So, for $i = 2$ we get ensembles produced by two (2) individual models like \textit{ARIMA\_Prophet}, \textit{ARIMA\_TBAT}, \textit{ARIMA\_N-BEATS} ... 
\textit{Prophet\_TBAT} etc. Accordingly, for $i = 3$, we get ensembles produced by three (3) individual models like  
\textit{ARIMA\_Prophet\_TBAT}, 
\textit{ARIMA\_Prophet\_N-BEATS} ... etc  and so on. Of course, only unique ensemble combinations are created, i.e,  duplicate ensembles like \textit{Prophet\_ARIMA} are not generated.

\section{Evaluation and Results}

In this section we first present the metrics used to evaluate our proposed approaches and afterwards we evaluate and compare the time series forecasting  methods with the machine learning regression algorithms for the same prediction problem. In the last section, the results of the two classification use cases are evaluated and discussed.

\subsection{Metrics}\label{metrics}
The following  metrics were  used to evaluate the results of the univariate time series forecasting and machine learning regression models :

\begin{equation}
MAE = \displaystyle\frac{1}{n}\sum_{t=1}^{n}|e_t|
\end{equation}

\begin{equation}
MAPE =\displaystyle\frac{100\%}{n}\sum_{t=1}^{n}\left |\frac{e_t}{y_t}\right|
\end{equation}

\begin{equation}
RMSE = \displaystyle\sqrt{\frac{1}{n}\sum_{t=1}^{n}e_t^2} 
\end{equation}

\begin{equation}
    MASE =  \displaystyle\frac{MAE}{MAE_{in-sample,naive}}
\end{equation}
\textit{where $e_t$ is the absolute error, $n$ is the sample size and $y_t$ is the actual value}

\vspace{\baselineskip}

The first three are quite common in the literature, while the last one is an interesting  metric that gives us an alternative baseline for determining the quality of our forecasts.

\begin{itemize}
    \item \textit{Mean Absolute Percentage Error (MAPE)}, expresses the average of the absolute percentage errors. While it is easily interpretable and  one of the most popular metrics, it has also  certain disadvantages \cite{MAKRIDAKIS1993527,RePEc:eee:intfor:v:8:y:1992:i:1:p:69-80} that lead us to utilizing  other metrics as well for a more reliable and complete evaluation of the models.
    \item \textit{Mean Absolute Error (MAE)}, is a data scale dependent metric which indicates  how big of an error we can expect from the forecast on average.
    \item \textit{Root Mean Squared Error (RMSE)}, is  the standard deviation of the residuals (prediction errors). It is a  data scale dependent metric  sensitive to outliers, putting  a heavier weight on larger errors.
    \item \textit{Mean Absolute Scaled Error (MASE)}, is a  measure of forecast accuracy which compares the model's  forecast against a naive benchmark method calculated in-sample. This measure is data scale independent, useful in cases where there are  different  scales in the data or values  which are negative or close to zero. In addition, it is easily interpretable: when $MASE < 1$, it implies  that the forecasts of the proposed method  perform, on average, better out-of-sample than  the in-sample one-step forecasts of  the naive method \cite{HYNDMAN2006679}.

As a benchmark reference, a simple naive model sets all forecasts to be the value of the last observation. That is, at the time $t$, the $k$-step-ahead naive forecast $F_{t+k}$ is ``predicted" with the observed value at time $t (y_t):$
\begin{equation}
F_{t+k} = y_t
\end{equation}

The naive static average model sets the forecasts to the average value using the  expanding window method and the moving  median and average models set the forecasts with the median and average values calculated with the rolling window  method.
\end{itemize}

\vspace{\baselineskip}

Concerning our classification approach, the evaluation metrics we use are the following:
\vspace{\baselineskip}

\begin{equation}
Accuracy = \displaystyle\frac{TP + TN}{TP + TN + FP + FN }  
\end{equation}

\begin{equation}
Precision  =\frac{TP}{TP + FP}
\end{equation}

\begin{equation}
Recall  = \displaystyle\frac{TP}{TP + FN } 
\end{equation}

\begin{equation}
\text{F-Measure} = \displaystyle 2 * \frac{\text{Precision * Recall}}{\text{Precision+Recall}}
\end{equation}
\textit{where $TP$, $TN$ are the True Positives and Negatives and $FP$, $FN$ the False Positives and Negatives predicted outcomes of the model}

\vspace{\baselineskip}

\begin{itemize}
    \item \textit{Accuracy},  is intuitively the overall fraction of predictions our model got right. However, it is not enough to evaluate our models using only this metric, especially for class-imbalanced datasets like ours. In this way, we examine the following metrics to get a more clear view of the model performance.
    \item \textit{Precision}, represents the  proportion of positive identifications that were actually correct
    \item \textit{Recall}, represents the proportion of the actual positives that were identified correctly
    \item \textit{F-measure},  is a measure of a test’s accuracy. It provides a way to combine both precision and recall into a single measure that captures both properties.
\end{itemize}

\subsection{Comparison between Forecasting and Regression Models}
In this part, the comparative performance of the regression algorithms and the univariate forecasting models is discussed. Our purpose is to examine which model category and which specific models perform better, what is the performance of the ensembles and what is the overall predictive ability for our use case of interest. We employ the aforementioned model categories with their respective individual models and ensembles, to make the prediction for our use case. We perform the training of the models with the 75\% of the data, while the evaluation is conducted with the rest 25\%, both sets of the original dataset. Due to time series indexing, this means that the most recent data  are used for the actual forecasts. To evaluate the overall performance in this use case we choose to present the following top five models produced per modeling category.

\vspace{0.5cm}
\textbf{Forecasting ensembles} :
\begin{enumerate}
    \item \textit{ARIMA, TBAT, NBEATS, TBATS}
    \item \textit{ARIMA, TBAT, NBEATS}
    \item \textit{TBAT, NBEATS, TBATS}
    \item \textit{Prophet, TBAT, NBEATS, TBATS}
    \item \textit{ARIMA, TBAT, TBATS}
\end{enumerate}

\textbf{Regression ensembles} :
\begin{enumerate}
   \item \textit{{AdaBoost, XGB, Decision Tree}}
   \item \textit{GBR, AdaBoost, XGB, Decision Tree}
   \item \textit{GBR, AdaBoost, Decision Tree}
   \item \textit{XGB, Extra Tree, Decision Tree}
   \item \textit{XGB, Decision Tree}
\end{enumerate}

In Tables ~\ref{table:results-forecasting} and ~\ref{table:results-regression} the detailed results of these best five models are presented per modeling category. We note that results for  MAE and RSME metric are expressed in minute scale. Considering both tables, we first point out the fact that ensembles make the top of the leader board as their resulting combined averaged forecasts are more accurate than those of the individual models and we also notice a comparable performance from both modeling categories. In particular, forecasting methods did slightly better in terms of relative errors (MAPE) but regression models were slightly better regarding the Mean Absolute and Root Mean Squared Errors. So, taking into account all these metrics, with each one having it's specific importance, we can conclude that in this use case the two modeling categories were equally competitive. In addition, we should also highlight the results of the MASE metric, i.e, the benchmark comparison against the four naive methods described in Section \ref{metrics}. We can tell that our proposed models produced in average more accurate forecasts than naive models. This is interesting, as it means that we can employ a meaningful model, at least in terms of implementation worthiness, comparing to a naive forecasting solution. 

Examining the winning ensembles more in depth, we observe on the one side, that TBAT algorithm  with its variation, TBATS, dominates in the forecasting ensembles. We also notice the model type variation in those winning ensembles, as we do not get ensembles of closely related models, but we witness the combinations of different algorithms, especially  those which include the N-BEATS deep neural architecture. On the other side, we observe that decision tree and secondly XGBoost, are core contributor models in each ensemble.

\begin{figure*}
\caption{Actual data and model performance  visualization}
\includegraphics[scale=0.75, width=\textwidth]{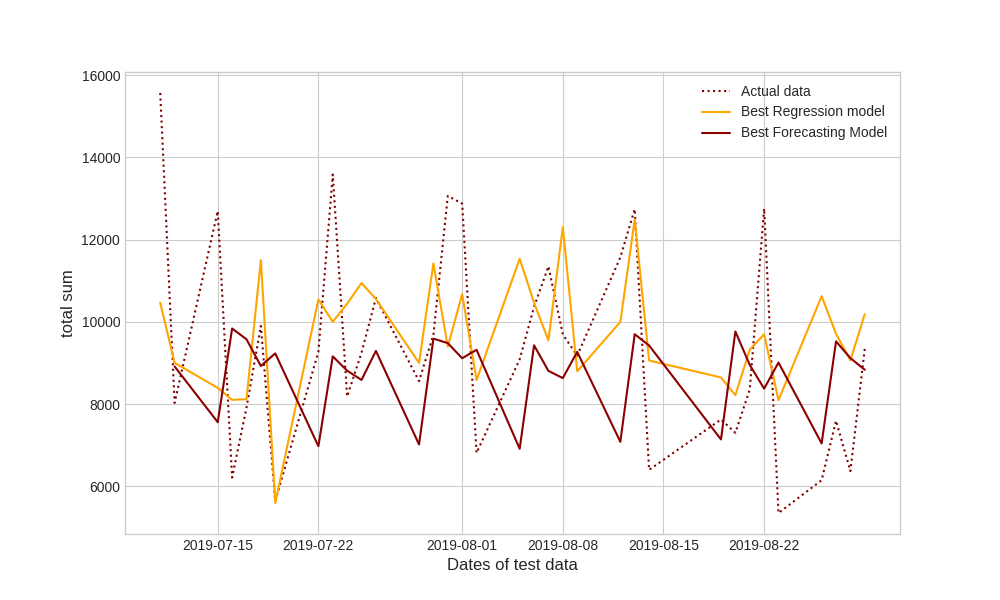}
\label{fig:1}
\end{figure*}

In Figure ~\ref{fig:1} we demonstrate a visualized comparison of the actual test data and the forecasts produced by the best model from the forecasting and regression category. The data set comprises of 34 daily data points ranging from 12th of July to 30th of August, as depicted  in the x-axis and the  total sum of minor\_stoppages expressed in seconds.

\begin{table*}
\centering
\caption{Results of the best five Forecasting Models}
\hspace*{1cm}\begin{tabularx}{\textwidth}{cccc|cccc @{\extracolsep{\fill}} ccp{6cm}}

    & & & & \multicolumn{4}{c}{MASE} \\

    Model & MAPE  & MAE & RMSE & moving\_mean & static\_mean & moving\_median & naive  \\
   
    1 & 19,765 \% & 32,536 & 42,399 & 0,926 & 0,882 & 0,920 & 0,664 \\
  
    2 & 19,833 \% & 33,026 & 43,255 & 0,940 & 0,895 & 0,934 & 0,674 \\ 
   
    3 & 20,007 \% & 33,808 & 44,282 & 0,962 & 0,916 & 0,956 & 0,690 \\ 
   
    4 & 20,135 \% & 33,165 & 43,435 & 0,944 & 0,899 & 0,938 & 0,677 \\ 
   
    5 & 20,156 \% & 33,489 & 42,568 & 0,953 & 0,908 & 0,947 & 0,683 \\ 
    \end{tabularx}
    \label{table:results-forecasting}
\end{table*}

\hspace*{2cm}\begin{table*}
\centering
\caption{Results of the best five Regression Models}
    
    \hspace*{1cm}\begin{tabularx}{\textwidth}{cccc|cccc @{\extracolsep{\fill}} ccp{6cm}}

    & & & & \multicolumn{4}{c}{MASE} \\
   
    Model & MAPE  & MAE & RMSE & moving\_mean & static\_mean & moving\_median & naive  \\
    1 & 20,753 \% & 31,085 & 37,988 & 0,890 & 0,842 & 0,866 & 0,632 \\

    2 & 20,984 \% & 31,727 & 38,368 & 0,909 & 0,860 & 0,884 & 0,645 \\ 

    3 & 21,017 \% & 31,466 & 37,879 & 0,901 & 0,853 & 0,877 & 0,639 \\ 

    4 & 21,080 \% & 31,458  & 38,553 & 0,901 & 0,852 & 0,877 & 0,639 \\ 

   5 & 21,252 \%  & 32,118 & 39,735 & 0,920 & 0,870 & 0,895 & 0,653 \\ 
    \end{tabularx}
    \label{table:results-regression}
\end{table*}

We notice the steady, seasonal alike prediction pattern of the forecasting ensemble which is close to the mean of the data. This is kind of expected, due to the underlying modeling process  of the individual models. In contrary, the regression model differentiates by producing higher and more custom forecasts, in some cases it even results interestingly in really high accuracy, like in the dates 16th, 27th of July and 14th of August.

Closing up, we acknowledge the decent performance of the models in regards to our agnostic modeling. In particular, the reframing of the time series data as a supervised  regression problem can be regarded successful, considering that the regression models achieved equal performance with the forecasting models, which are tailored for specific univariate time series problems. Research wise, we could extend the experiments in the future with the collection of additional data, in order to have a more holistic view of the trends, seasonality and other patterns and thus, achieve better forecasting accuracy. We could also try working on different subsets of the  data to examine the  performance  of the models in relation to  the dataset size. Lastly, it would be interesting to further experiment with ensembles between models that have been trained in different subsets of the data as well as ensembles between the forecasting and regression models. 

\subsection{Classification Use Cases}
In this part, we present the results of the Random Forest classifier for the two aforementioned prediction use cases. For simplifying our findings  evaluation, we can regard for each use case accordingly, the positive class as an alarm generation towards the industrial operators and the negative class as an ignore state, assuming that the predictions were incorporated in a real time predictive maintenance system. For both of them we perform the training and evaluation of the models in the same way as in our regression approach, applying a  75\% / 25\% percentage split between the training and test sets. This is also approximately the class distribution for both cases,  75\% and 25 \% for the negative and positive class respectively. For each use case we present the generated confusion matrix\footnote{Also known as an error matrix, it is a specific table layout that allows visualization of the performance of an algorithm, typically a supervised learning one (in unsupervised learning it is usually called a matching matrix). Each column of the matrix represents the instances in a predicted class while each row represents the instances in an actual class} (Tables ~\ref{table:confusion-matrix-case1} and ~\ref{table:confusion-matrix-case2}) which consists of the predicted True \& False Positives and True \& False Negatives, based on which the classifications metrics are derived in Tables  ~\ref{table:metrics-case1} and ~\ref{table:metrics-case2}.

\subsubsection{Minor Stoppage Duration Exceeding}

In the dataset we treat the classes as follows: positive (1) is the class describing that  the total sum of minor stoppages will exceed the duration of 10 minutes in the next hour and negative (0)  is the class about the opposite. The dataset focuses on the period of April 2019 till July 2019,  which is the operational peak of the production line and the packing machine. Considering tables~\ref{table:confusion-matrix-case1} and~\ref{table:metrics-case1}, we observe that the classifier  detects only 11 \% of the actual alarms it should detect, which is in fact low. However, we notice an optimistic result, the almost absolute identification of the ignore class, with 99\% Recall for ignore class and 78\% Precision for the alarm class, meaning that the classifier at least would not signal many false alarm warnings. This is indeed a valuable predictive ability, otherwise, producing frequent  false alarms may lead to the industrial operator's distraction. Or even worse, inaccurate predictions can make them take wrong preventive and corrective decisions for the production line and the investigated machine.

\begin{table}[!htbp]
\caption{Confusion matrix of Minor Stoppage Duration Exceeding case}
    \centering
\makegapedcells
\begin{tabular}{cc|cc}
\multicolumn{2}{c}{}
            &   \multicolumn{2}{c}{Predicted} \\
    &       &  ignore  & alarm               \\ 
    \cline{2-4}
\multirow{2}{*}{\rotatebox[origin=c]{90}{Actual}}
    & ignore   & 369   & 4                 \\
    & alarm    & 113    & 14                \\ 
    \cline{2-4}
\end{tabular}
    \label{table:confusion-matrix-case1}
\end{table}
   
\begin{table}[!htbp]
 \caption{Performance of Minor Stoppage Duration Exceeding case}
    \begin{tabular}{|c|c|c|c|c|}
    \hline
   Accuracy & Class & Precision & Recall & F-measure \\
    \hline
 \multirow{2}{*}{77 \%} & ignore & 77 \% & 99 \% & 86 \% \\
   \cline{2-5}
   & alarm & 78 \% & 11 \% & 19 \%   \\
    \hline
    \end{tabular}
    \label{table:metrics-case1}
\end{table}

\subsubsection{Breakdown Occurrence}
This use case is modeled between January 2019 and September 2019, the period containing the most observations in our dataset. In this way, our aim is to gather in the input data as many breakdown events and relevant information around their occurrence as possible, in order to increase the chances of a successful prediction. We treat as positive (1) class, the event of a breakdown occurence in the next hour and as negative (0) class, the exact opposite event. Reviewing the results in Tables \ref{table:confusion-matrix-case2} and ~\ref{table:metrics-case2}, we notice that our classifier performed better in this case than the previous one, in the prediction of the actual occurrences  of the main class of interest (in this case, "alarm" for a possible breakdown). The improvement was by a factor of 10 \%, reaching a Recall of 21\% (compared to 11\% of the previous case). However, this was caused, by the reduced precision, i.e, the reduced ability of the model to distinguish clearly between the actual and false alarms and so the alarm class has only, in average, 51\% chance to be predicted correctly (compared to 78\% of the previous case). This is a usual phenomenon in machine learning, also known as precision and recall trade-off, in which improving  the former metric might worsen the latter and vice versa. The predictive power  for the ignore (negative) class is slightly  worse than the first use case, as precision dropped from 77\% to 75\%, but it is still a relatively good sign that  the model can at least perform well for one of the metrics.

\begin{table}[!htbp]
\caption{Confusion matrix of Breakdown Occurrence case}
    \centering
\makegapedcells
\begin{tabular}{cc|cc}
\multicolumn{2}{c}{}
            &   \multicolumn{2}{c}{Predicted} \\
    &       &  ignore  & alarm                \\ 
    \cline{2-4}
\multirow{2}{*}{\rotatebox[origin=c]{90}{Actual}}
    & ignore   & 804   & 69                 \\
    & alarm    & 274    & 72                \\ 
    \cline{2-4}
\end{tabular}

    \label{table:confusion-matrix-case2}
\end{table}

\begin{table}[!htbp]
\caption{Performance of Breakdown Occurrence case}
    \begin{tabular}{|c|c|c|c|c|}
    \hline
    Accuracy & Class & Precision & Recall & F-measure \\
    \hline
 \multirow{2}{*}{72 \%}  & ignore & 75 \% & 92 \% & 82 \% \\
   \cline{2-5}
   & alarm & 51 \% & 21 \% & 30 \%   \\
    \hline
    \end{tabular}
    
    \label{table:metrics-case2}
\end{table}



\begin{figure}[h!] 
    \centering
    
    \begin{subfigure}[t]{\columnwidth}
        \includegraphics[width=\columnwidth]{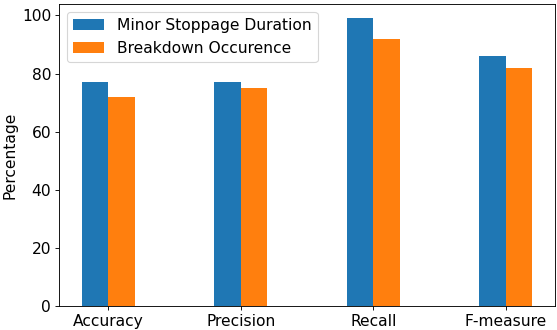}
        \label{fig:water-DT}
    \end{subfigure}
    \begin{subfigure}[t]{\columnwidth}
           \includegraphics[width=\columnwidth]{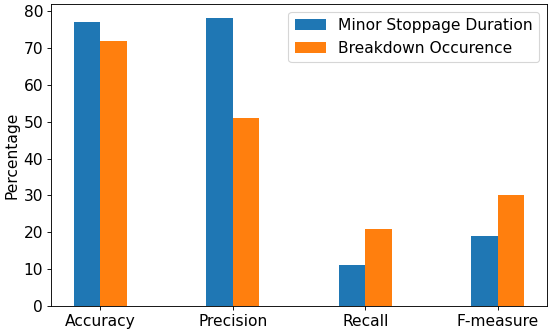}
        \label{fig:water-TPOT}
    \end{subfigure}
    
    \caption{Performance of Random Forest in both classification use cases, per class. The figure at the top depicts the performance for the ignore (negative) class, while the bottom one is for the alarm (positive) class.}
    \label{fig:performance-metrics}
\end{figure}

In Figure \ref{fig:performance-metrics}, the performance of Random Forest is provided to visualize the differences in each metric, per use case and prediction class. The figure at the top depicts the performance for the ignore (negative) class, while the bottom one is for the alarm (positive) class.

Given these results for both use cases and looking further into the F-measure values, we can sum up that our feature-based time series classification approach achieved an overall moderate, weak performance in the alarm (positive) class. On the contrary, it reveals interesting results for the ignore (negative) class, which seems to be predicted in a quite accurate way. The tests indicate that we miss essential information in order to achieve high accuracy in the prediction of alarms cases and this is not to our surprise, as we do not monitor the different factors that may cause a stoppage and in general disturbances in the production line. This means that future research activities will include the collection of those exogenous factors and the further modeling experimentation of the aforementioned use cases. Nonetheless, our agnostic methodology so far can be still regarded as decent work, considering the results produced while having only one signal at our disposal. In this way, the regression and classification approaches we developed can be utilized as a baseline work for these future activities.

\section{Conclusion}\label{sec:sec10}
In this work, we investigated the prediction of industrial packing machine stoppages by applying the approaches of traditional univariate time series forecasting and supervised machine learning. Our highest forecasting and classification accuracy of 80\% and  77\% respectively does not imply a perfect, state-of-the-art performance but we can still regard these results as promising considering our agnostic methodology, indicating that it is feasible, at least to a certain extent, to model the machine's behavior based only on its raw past operational state measurements. We acknowledge that the presence of noise and random fluctuations in the data as well as the lack of strong seasonalities, clear structural patterns, and most importantly the absence of the stoppage factors weaken the forecasting and predictive ability of the models. Therefore, we note that better results can be potentially achieved with the use of more production line data. Such data could be derived from additional sensors and operational variables, fault or warning signals from the same machine as well as from other machines which operate in the same line. This is due to the fact that such data affect the general flow of the entire packaging line and consequently the operation of the packing machine. Hence, it is clear that prospects for future research include the collection of such data and use it in our modeling. In addition to the aforementioned update and enrichment of our data pool, our proposed methodologies can act as a reference for future enhancements with more suitable predictive models and techniques (such as deep learning, feature selection methods, classification ensembles, and class imbalance handling) that can eventually lead to a more robust and accurate prediction framework.\ITUpar

\section*{Acknowledgement}
\label{sec:ackn}
This research was supported by (i) the ``Andreas Mentzelopoulos Foundation'', (ii) the CNR Short Term Mobility (STM) program 2019, and (iii) the European Union's Horizon Europe research and innovation programme RE4DY (European Data as a PRoduct Value Ecosystems for Resilient Factory 4.0 Product and ProDuction ContinuitY and Sustainability) under grant agreement No 101058384.\ITUpar

\bibliographystyle{IEEEtran}
\bibliography{references}

\section*{Authors}
\label{sec:auth}

\begin{wrapfigure}{l}{0.32\columnwidth} 
    \vspace{-.1in}
    \includegraphics[width=0.39\columnwidth, height=3cm]{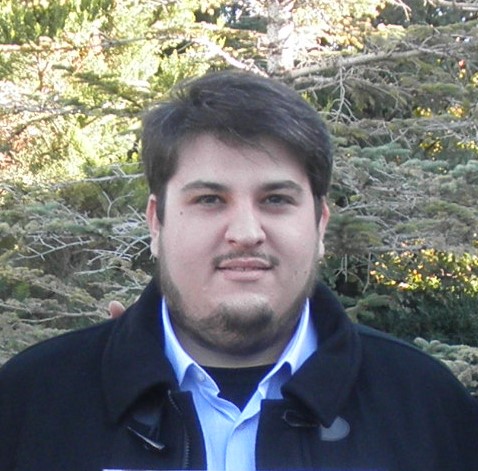} 
\end{wrapfigure}\textbf{Gabriel Filios} is a Postdoctoral Researcher at the Computer Engineering \& Informatics department in the University of Patras, Greece. His research interests include Wireless Sensor Networks, Energy Efficiency in smart Buildings and Industries, Feature Extraction for Human Activity, Crowdsensing Systems and Privacy in IoT. He has co-authored several publications in international refereed conferences (IEEE ICC, ACM MobiWac, IEEE WF-IoT) and has participated in several EU Projects (HOBNET, IoT Lab, PrivacyFlag, SAFESTRIP).\ITUpar

\begin{wrapfigure}{l}{0.32\columnwidth} 
    \vspace{-.1in}
    \includegraphics[width=0.39\columnwidth, height=3cm]{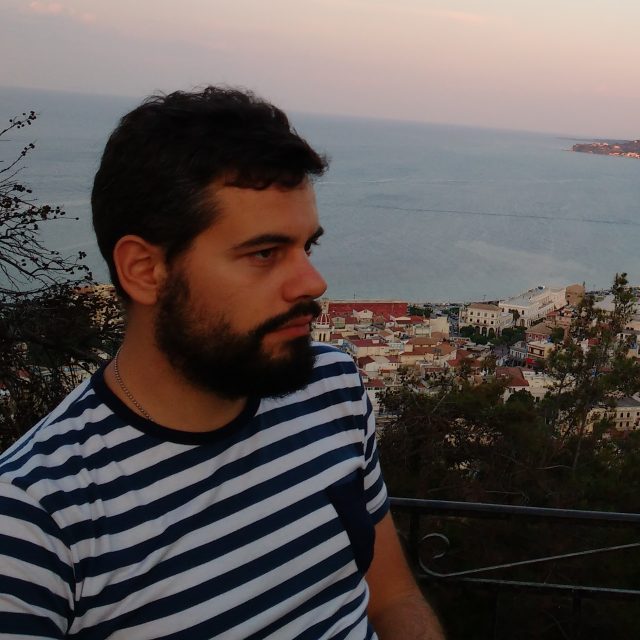} 
\end{wrapfigure}\textbf{Ioannis Katsidimas} is a Postdoctoral Researcher at the Computer Engineering and Informatics Department of Patras University, Greece. His research interests include Wireless Power Transfer algorithms in Ad hoc Communication Networks, Wireless Sensor Networks, Internet of Things. He has participated in several EU Projects (IoT Lab, PrivacyFlag, SAFESTRIP).\ITUpar

\begin{wrapfigure}{l}{0.32\columnwidth} 
    \vspace{-.1in}
    \includegraphics[width=0.39\columnwidth, height=3cm]{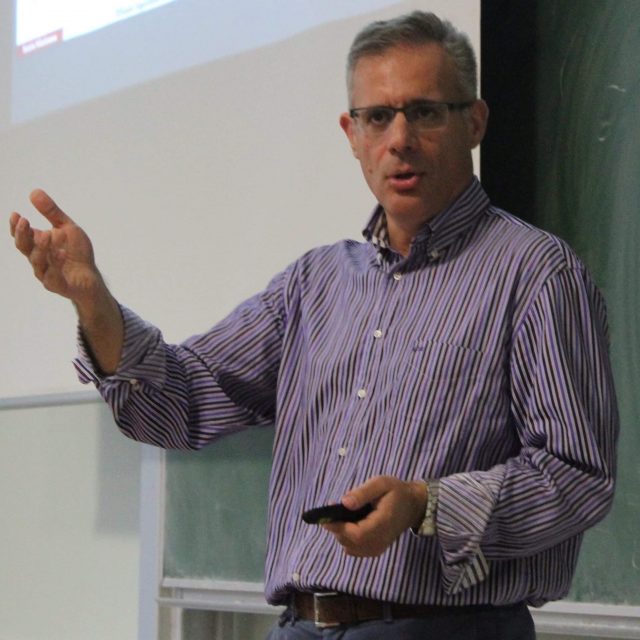} 
\end{wrapfigure}\textbf{Sotiris Nikoletseas} is a Full Professor at the Computer Engineering and Informatics Department of Patras University, Greece. His research interests include algorithmic aspects of wireless sensor networks and the Internet of Things (IoT), wireless energy transfer protocols, probabilistic techniques and random graphs, average case analysis and probabilistic algorithms, algorithmic engineering. He has coauthored over 300 publications in international Journals and refereed Conferences, 3 Books (on the Probabilistic Method, on theoretical aspects of sensor networks, on wireless power) and 30 Invited Chapters in Books by major publishers.\ITUpar

\begin{wrapfigure}{l}{0.32\columnwidth} 
    \vspace{-.1in}
    \includegraphics[width=0.39\columnwidth, height=3cm]{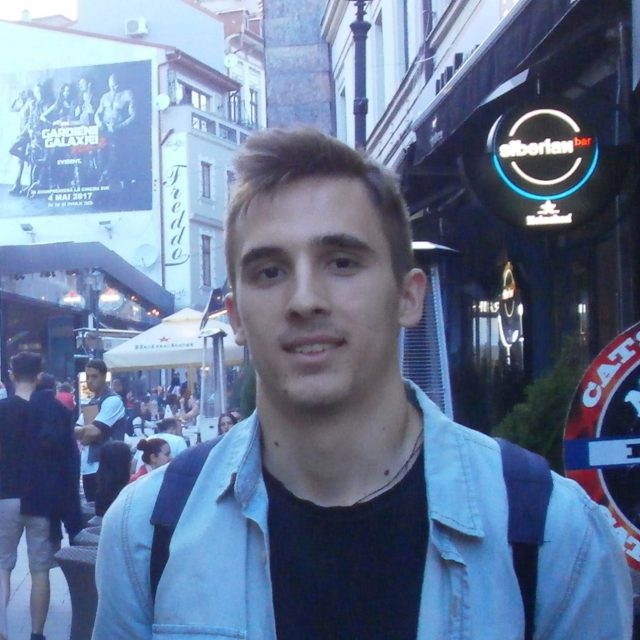} 
\end{wrapfigure}\textbf{Stefanos H. Panagiotou} is a PhD Candidate at the Computer Engineering and Informatics Department of Patras University, Greece. His research is being held under scholarship funding from University of Patras and his interests include machine learning applications in production lines, TinyML and AI-driven digital twins for structural health monitoring. \ITUpar

\begin{wrapfigure}{l}{0.32\columnwidth} 
    \vspace{-.1in}
    \includegraphics[width=0.39\columnwidth, height=3cm]{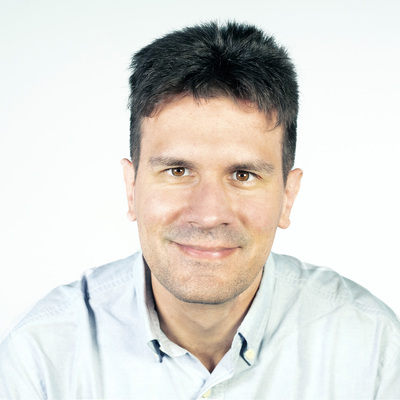} 
\end{wrapfigure}\textbf{Theofanis P. Raptis} received the Ph.D. degree from the University of Patras, Greece. He is currently a Research Scientist with the National Research Council, Italy. He has published in journals, conference proceedings, and books, more than 70 articles on industrial networks, wirelessly powered networks, Internet of Things testbeds, and platforms. He is also regularly involved in international IEEE and ACM sponsored conference and workshop organization committees, in the areas of networks, computing, and communications. He has been serving as an Associate Editor for the IEEE Access and IET Networks journals.\ITUpar
\end{document}